# Discourse Processing of Dialogues with Multiple Threads

Carolyn Penstein Rosé[†], Barbara Di Eugenio[†], Lori S. Levin[†],
Carol Van Ess-Dykema[‡]

[†] Computational Linguistics Program
Carnegie Mellon University
Pittsburgh, PA, 15213
{cprose,dieugeni}@lcl.cmu.edu
lsl@cs.cmu.edu

[‡] Department of Defense
Mail stop: R525
9800 Savage Road
Ft. George G. Meade, MD 20755-6000
cjvanes@afterlife.ncsc.mil

## Abstract

In this paper we will present our ongoing work on a plan-based discourse processor developed in the context of the Enthusiast Spanish to English translation system as part of the JANUS multi-lingual speech-to-speech translation system. We will demonstrate that theories of discourse which postulate a strict tree structure of discourse on either the intentional or attentional level are not totally adequate for handling spontaneous dialogues. We will present our extension to this approach along with its implementation in our plan-based discourse processor. We will demonstrate that the implementation of our approach outperforms an implementation based on the strict tree structure approach.

## 1 Introduction

In this paper we will present our ongoing work on a plan-based discourse processor developed in the context of the Enthusiast Spanish to English translation system (Suhm et al. 1994) as part of the JANUS multi-lingual speech-to-speech translation system. The focus of the work reported here has been to draw upon techniques developed recently in the computational discourse processing community (Lambert 1994; Lambert 1993; Hinkelman 1990), developing a discourse processor flexible enough to cover a large corpus of spontaneous dialogues in which two speakers attempt to schedule a meeting.

There are two main contributions of the work we will discuss in this paper. From a theoretical standpoint, we will demonstrate that theories which postulate a strict tree structure of discourse (henceforth, Tree Structure Theory, or TST) on either the intentional level or the attentional level (Grosz and Sidner 1986) are not totally adequate for covering spontaneous dialogues, particularly negotiation dialogues which are composed of multiple threads. These are negotiation dialogues in which multiple propositions are negotiated in parallel. We will discuss our proposed extension to TST which handles these structures in a perspicuous manner. From a practical standpoint, our second contribution will be a description of our implemented discourse processor which makes use of this extension of TST, taking as input the imperfect result of parsing these spontaneous dialogues.

We will also present a comparison of the performance of two versions of our discourse processor, one based on strict TST, and one with our extended version of TST, demonstrating that our extension of TST yields an improvement in performance on spontaneous scheduling dialogues.

A strength of our discourse processor is that because it was designed to take a language-independent meaning representation (interlingua) as its input, it runs without modification on either English or Spanish input. Development of our discourse processor was based on a corpus of 20 spontaneous Spanish scheduling dialogues containing a total of 630 sentences. Although development and initial testing of the discourse processor was done with Spanish dialogues, the theoretical work on the model as well as the evaluation presented in this paper was done with spontaneous English dialogues.

In section 2, we will argue that our proposed extension to Standard TST is necessary for making correct predictions about patterns of referring expressions found in dialogues where multiple alternatives are argued in parallel. In section 3 we will present our implementation of Extended TST. Finally, in section 4 we will present an evaluation of the performance of our discourse processor with Extended TST compared to its performance using Standard TST.

## 2 Discourse Structure

Our discourse model is based on an analysis of naturally occurring scheduling dialogues. Figures 1 and 2 contain examples which are adapted from naturally occurring scheduling dialogues. These examples contain the sorts of phenomena we have found in our corpus but have been been simplified for the

(1) S1: We need to set up a schedule for the meeting.

(2)     How does your schedule look for next week?
(3) S2: Well, Monday and Tuesday both mornings are good.
(4)     Wednesday afternoon is good also.
(5) S1: It looks like it will have to be Thursday then.
(6)     Or Friday would also possibly work.
(7)     Do you have time between twelve and two on Thursday?
(8)     Or do you think sometime Friday afternoon you could meet?
(9) S2: No.
(10)    Thursday I have a class.
(11)    And Friday is really tight for me.
(12)    How is the next week?
(13)    If all else fails there is always video conferencing.
(14) S1: Monday, Tuesday, and Wednesday I am out of town.
(15)    But Thursday and Friday are both good.
(16)    How about Thursday at twelve?
(17) S2: Sounds good.

(18)    See you then.

Figure 1: **Example of Deliberating Over A Meeting Time**

purpose of making our argument easy to follow. Notice that in both of these examples, the speakers negotiate over multiple alternatives in parallel.

We challenge an assumption underlying the best known theories of discourse structure (Grosz and Sidner 1986; Scha and Polanyi 1988; Polanyi 1988; Mann and Thompson 1986), namely that discourse has a recursive, tree-like structure. Webber (1991) points out that Attentional State[1] is modeled equivalently as a stack, as in Grosz and Sidner's approach, or by constraining the current discourse segment to attach on the rightmost frontier of the discourse structure, as in Polanyi and Scha's approach. This is because *attaching* a leaf node corresponds to *pushing* a new element on the stack; *adjoining* a node $D_i$ to a node $D_j$ corresponds to *popping* all the stack elements through the one corresponding to $D_j$ and *pushing* $D_i$ on the stack. Grosz and Sider (1986), and more recently Lochbaum (1994), do not formally constrain their intentional structure to a strict tree structure, but they effectively impose this limitation in cases where an anaphoric link must be made between an expression inside of the current discourse segment and an entity evoked in a different segment. If the expression can only refer to an entity on the stack, then the discourse segment purpose[2] of the current discourse segment must be attached to the rightmost frontier of the intentional structure. Otherwise the entity which the expression refers to would have already been popped from the stack by the time the reference would need to be resolved.

We develop our theory of discourse structure in the spirit of (Grosz and Sidner 1986) which has played an influential role in the analysis of discourse entity saliency and in the development of dialogue processing systems. Before we make our argument, we will argue for our approach to discourse segmentation. In a recent extension to Grosz and Sidner's original theory, described in (Lochbaum 1994), each discourse segment purpose corresponds to a partial or full shared plan[3] (Grosz and Kraus 1993). These discourse segment purposes are expressed in terms of the two intention operators described in (Grosz and Kraus 1993), namely *Int.To* which represents an agent's intention to perform some action and

---

[1] Attentional State is the representation which is used for computing which discourse entities are most salient.

[2] A discourse segment purpose denotes the goal which the speaker(s) attempt to accomplish in engaging in the associated segment of talk.

[3] A Shared Plan is a plan which a group of two or more participants intend to accomplish together.

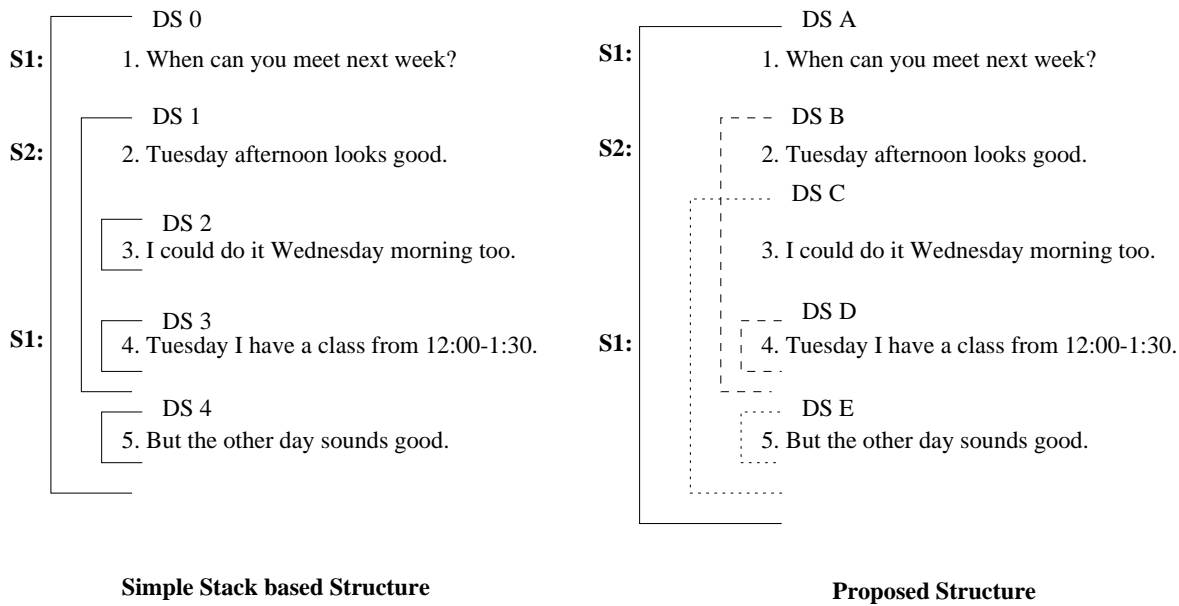

Figure 2: **Sample Analysis**

*Int.That* which represents an agent's intention that some proposition hold. Potential intentions are used to account for an agent's process of weighing different means for accomplishing an action he is committed to performing (Bratman, Israel, & Pollack 1988). These potential intentions, *Pot.Int.To* and *Pot.Int.That*, are not discourse segment purposes in Lochbaum's theory since they cannot form the basis for a shared plan having not been decided upon yet and being associated with only one agent. It is not until they have been decided upon that they become *Int.To*'s and *Int.That*'s which can then become discourse segment purposes. We argue that potential intentions must be able to be discourse segment purposes.

Potential intentions are expressed within portions of dialogues where speakers negotiate over how to accomplish a task which they are committed to completing together. For example, deliberation over how to accomplish a shared plan can be represented as an expression of multiple *Pot.Int.To*'s and *Pot.Int.That*'s, each corresponding to different alternatives. As we understand Lochbaum's theory, for each factor distinguishing these alternatives, the potential intentions are all discussed inside of a single discourse segment whose purpose is to explore the options so that the decision can be made.

The stipulation that *Int.To*'s and *Int.That*'s can be discourse segment purposes but *Pot.Int.To*'s and *Pot.Int.That*'s cannot has a major impact on the analysis of scheduling dialogues such as the one in Figure 1 since the majority of the exchanges in scheduling dialogues are devoted to deliberating over which date and at which time to schedule a meeting. This would seem to leave all of the deliberation over meeting times within a single monolithic discourse segment, leaving the vast majority of the dialogue with no segmentation. As a result, we are left with the question of how to account for shifts in focus which seem to occur within the deliberation segment as evidenced by the types of pronominal references which occur within it. For example, in the dialogue presented in Figure 1, how would it be possible to account for the differences in interpretation of "Monday" and "Tuesday" in (3) with "Monday" and "Tuesday" in (14)? It cannot simply be a matter of immediate focus since the week is never mentioned in (13). And there are no semantic clues in the sentences themselves to let the hearer know which week is intended. Either there is some sort of structure in this segment more fine grained than would be obtained if *Pot.Int.To*'s and *Pot.Int.That*'s cannot be discourse segment purposes, or another mechanism must be proposed to account for the shift in focus which occurs within the single segment. We argue that rather than propose an additional mechanism, it is more perspicuous to lift the restriction that *Pot.Int.To*'s and *Pot.Int.That*'s cannot be discourse segment purposes. In our approach a separate discourse segment is allocated for every potential plan discussed in the dialogue, one corresponding to each parallel potential intention expressed.

Assuming that potential intentions form the basis for discourse segment purposes just as intentions do, we present two alternative analyses for an example dialogue in Figure 2. The one on the left is the one which would be obtained if Attentional State were modeled as a stack. It has two shortcomings. The first is that the suggestion for meeting on Wednesday in DS 2 is treated like an interruption. Its focus space is pushed onto the stack and then popped off when the focus space for the response to the suggestion for Tuesday in DS 3 is pushed[4]. Clearly, this suggestion is not an interruption however. Furthermore, since the focus space for DS 2 is popped off when the focus space for DS 4 is pushed on, Wednesday is nowhere on the focus stack when "the other day", from sentence 5, must be resolved. The only time expression on the focus stack at that point would be "next week". But clearly this expression refers to Wednesday. So the other problem is that it makes it impossible to resolve anaphoric referring expressions adequately in the case where there are multiple threads, as in the case of parallel suggestions negotiated at once.

We approach this problem by modeling Attentional State as a graph structured stack rather than as a simple stack. A graph structured stack is a stack which can have multiple top elements at any point. Because it is possible to maintain more than one top element, it is possible to separate multiple threads in discourse by allowing the stack to branch out, keeping one branch for each thread, with the one most recently referred to more strongly in focus than the others. The analysis on the right hand side of Figure 2 shows the two branches in different patterns. In this case, it is possible to resolve the reference for "the other day" since it would still be on the stack when the reference would need to be resolved. Implications of this model of Attentional State are explored more fully in (Rosé 1995).

## 3 Discourse Processing

We evaluated the effectiveness of our theory of discourse structure in the context of our implemented discourse processor which is part of the Enthusiast Speech translation system. Traditionally machine translation systems have processed sentences in isolation. Recently, however, beginning with work at ATR, there has been an interest in making use of discourse information in machine translation. In (Iida and Arita 1990; Kogura et al. 1990), researchers at ATR advocate an approach to machine translation called illocutionary act based translation, arguing that equivalent sentence forms do not necessarily carry the same illocutionary force between languages. Our implementation is described more fully in (Rosé 1994). See Figure 4 for the discourse rep-

---

[4]Alternatively, DS 2 could not be treated like an interruption, in which case DS 1 would be popped before DS 2 would be pushed. The result would be the same. DS 2 would be popped before DS 3 would be pushed.

```
((when
    ((frame *simple-time)
     (day-of-week wednesday)
     (time-of-day morning)))
 (a-speech-act
    (*multiple* *suggest *accept))
 (who
    ((frame *i)))
 (frame *free)
 (sentence-type *state)))
```

**Sentence:** I could do it Wednesday morning too.

Figure 3: **Sample Interlingua Representation with Possible Speech Acts Noted**

resentation our discourse processor obtains for the example dialogue in Figure 2. Note that although a complete tripartite structure (Lambert 1993) is computed, only the discourse level is displayed here.

Development of our discourse processor was based on a corpus of 20 spontaneous Spanish scheduling dialogues containing a total of 630 sentences. These dialogues were transcribed and then parsed with the GLR* skipping parser (Lavie and Tomita 1993). The resulting interlingua structures (See Figure 3 for an example) were then processed by a set of matching rules which assigned a set of possible speech acts based on the interlingua representation returned by the parser similar to those described in (Hinkelman 1990). Notice that the list of possible speech acts resulting from the pattern matching process are inserted in the a-speech-act slot ('a' for ambiguous). It is the structure resulting from this pattern matching process which forms the input to the discourse processor. Our goals for the discourse processor include recognizing speech acts and resolving ellipsis and anaphora. In this paper we focus on the task of selecting the correct speech act.

Our discourse processor is an extension of Lambert's implementation (Lambert 1994; Lambert 1993; Lambert and Carberry 1991). We have chosen to pattern our discourse processor after Lambert's recent work because of its relatively broad coverage in comparison with other computational discourse models and because of the way it represents relationships between sentences, making it possible to recognize actions expressed over multiple sentences. We have left out aspects of Lambert's model which are too knowledge intensive to get the kind of coverage we need. We have also extended the set of structures recognized on the discourse level in order to identify speech acts such as Suggest, Accept, and Reject which are common in negotiation discourse. There are a total of thirteen possible speech acts which we identify with our discourse processor. See Figure 5 for a complete list.

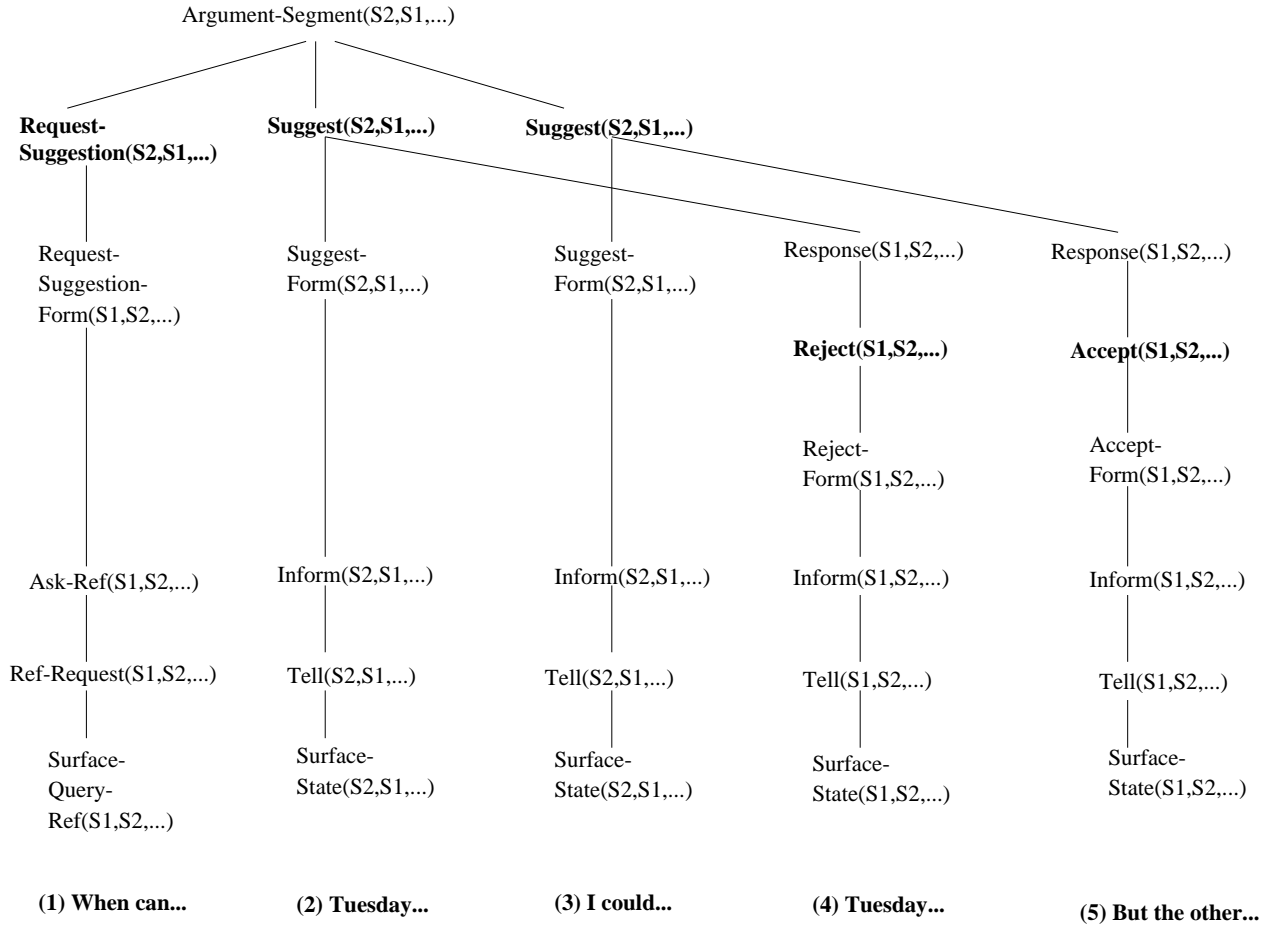

Figure 4: **Sample Discourse Structure**

It is commonly impossible to tell out of context which speech act might be performed by some utterances since without the disambiguating context they could perform multiple speech acts. For example, "I'm free Tuesday." could be either a Suggest or an Accept. "Tuesday I have a class." could be a State-Constraint or a Reject. And "So we can meet Tuesday at 5:00." could be a Suggest or a Confirm-Appointment. That is why it is important to construct a discourse model which makes it possible to make use of contextual information for the purpose of disambiguating.

Some speech acts have weaker forms associated with them in our model. Weaker and stronger forms very roughly correspond to direct and indirect speech acts. Because every suggestion, rejection, acceptance, or appointment confirmation is also giving information about the schedule of the speaker, State-Constraint is considered to be a weaker form of Suggest, Reject, Accept, and Confirm-Appointment.

Also, since every acceptance expressed as "yes" is also an affirmative answer, Affirm is considered to be a weaker form of Accept. Likewise Negate is considered a weaker form of Reject. This will come into play in the next section when we discuss our evaluation.

When the discourse processor computes a chain of inference for the current input sentence, it attaches it to the current plan tree. Where it attaches determines which speech act is assigned to the input sentence. For example, notice than in Figure 4, because sentences 4 and 5 attach as responses, they are assigned speech acts which are responses (i.e. either Accept or Reject). Since sentence 4 chains up to an instantiation of the Response operator from an instantiation of the Reject operator, it is assigned the speech act Reject. Similarly, sentence 5 chains up to an instantiation of the Response operator from an instantiation of the Accept operator, sentence 5 is assigned the speech act Accept. After the discourse processor attaches the current sentence to the plan tree thereby selecting the correct speech act in context, it inserts the correct speech act in the speech-act slot in the interlingua structure. Some speech acts are not recognized by attaching them to the previous plan tree. These are speech acts such as Suggest which are not responses to previous speech acts. These are recognized in cases where the plan inferencer chooses not to attach the current inference chain to the previous plan tree.

| Speech Act | Example |
| --- | --- |
| Opening | Hi, Cindy. |
| Closing | See you then. |
| Suggest | Are you free on the morning of the eighth? |
| Reject | Tuesday I have a class. |
| Accept | Thursday I'm free the whole day. |
| State-Constraint | This week looks pretty busy for me. |
| Confirm-Appointment | So Wednesday at 3:00 then? |
| Negate | no. |
| Affirm | yes. |
| Request-Response | What do you think? |
| Request-Suggestion | What looks good for you? |
| Request-Clarification | What did you say about Wednesday? |
| Request-Confirmation | You said Monday was free? |

Figure 5: **Speech Acts covered by the system**

When the chain of inference for the current sentence is attached to the plan tree, not only is the speech act selected, but the meaning representation for the current sentence is augmented from context. Currently we have only a limited version of this process implemented, namely one which augments the time expressions between previous time expressions and current time expressions. For example, consider the case where Tuesday, April eleventh has been suggested, and then the response only makes reference to Tuesday. When the response is attached to the suggestion, the rest of the time expression can be filled in.

The decision of which chain of inference to select and where to attach the chosen chain, if anywhere, is made by the focusing heuristic which is a version of the one described in (Lambert 1993) which has been modified to reflect our theory of discourse structure. In Lambert's model, the focus stack is represented implicitly in the rightmost frontier of the plan tree called the active path. In order to have a focus stack which can branch out like a graph structured stack in this framework, we have extended Lambert's plan operator formalism to include annotations on the actions in the body of decomposition plan operators which indicate whether that action should appear 0 or 1 times, 0 or more times, 1 or more times, or exactly 1 time. When an attachment to the active path is attempted, a regular expression evaluator checks to see that it is acceptable to make that attachment according to the annotations in the plan operator of which this new action would become a child. If an action on the active path is a repeating action, rather than only the rightmost instance being included on the active path, all adjacent instances of this repeating action would be included. For example, in Figure 4, after sentence 3, not only is the second, rightmost suggestion in focus, along with its corresponding inference chain, but both suggestions are in focus, with the rightmost one being slightly more accessible than the previous one. So when the first response is processed, it can attach to the first suggestion. And when the second response is processed, it can be attached to the second suggestion. Both suggestions remain in focus as long as the node which immediately dominates the parallel suggestions is on the rightmost frontier of the plan tree. Our version of Lambert's focusing heuristic is described in more detail in (Rosé 1994).

## 4  Evaluation

The evaluation was conducted on a corpus of 8 previously unseen spontaneous English dialogues containing a total of 223 sentences. Because spoken language is imperfect to begin with, and because the parsing process is imperfect as well, the input to the discourse processor was far from ideal. We are encouraged by the promising results presented in figure 6, indicating both that it is possible to successfully process a good measure of spontaneous dialogues in a restricted domain with current technology,[5] and that our extension of TST yields an improvement in performance.

The performance of the discourse processor was evaluated primarily on its ability to assign the correct speech act to each sentence. We are not claiming that speech act recognition is the best way to evaluate the validity of a theory of discourse, but because speech act recognition is the main aspect of the discourse processor which we have implemented, and because recognizing the discourse structure is part of the process of identifying the correct speech act, we believe it was the best way to evaluate the difference between the two different focusing mechanisms in our implementation at this time. Prior to the evaluation, the dialogues were analyzed by hand and sentences were assigned their correct speech act for comparison with those eventually selected by the discourse processor. Because the speech acts for the test dialogues were coded by one of the authors and we do not have reliability statistics for this encoding, we would draw the attention of the readers more to the difference in performance between the two focusing mechanisms rather than to the absolute performance in either case.

---

[5]It should be noted that we do not claim to have solved the problem of discourse processing of spontaneous dialogues. Our approach is coursely grained and leaves much room for future development in every respect.

| Version | Good | Acceptable | Incorrect |
|---|---|---|---|
| **Extended TST** | 171 total (77%) 144 based on plan-inference | 27 total (12%) 22 based on plan inference | 25 total (11%) 20 based on plan inference |
| **Standard TST** | 161 total (72%) 116 based on plan inference | 33 total (15%) 25 based on plan inference | 28 total (13%) 23 based on plan inference |

Figure 6: **Performance Evaluation Results**

For each sentence, if the correct speech act, or either of two equally preferred best speech acts were recognized, it was counted as correct. If a weaker form of a correct speech act was recognized, it was counted as acceptable. See the previous section for more discussion about weaker forms of speech acts. Note that if a stronger form is recognized when only the weaker one is correct, it is counted as wrong. And all other cases were counted as wrong as well, for example recognizing a suggestion as an acceptance.

In each category, the number of speech acts determined based on plan inference is noted. In some cases, the discourse processor is not able to assign a speech act based on plan inference. In these cases, it randomly picks a speech act from the list of possible speech acts returned from the matching rules. The number of sentences which the discourse processor was able to assign a speech act based on plan inference increases from 164 (74%) with Standard TST to 186 (83%) with Extended TST. As Figure 6 indicates, in many of these cases, the discourse processor guesses correctly. It should be noted that although the correct speech act can be identified without plan inference in many cases, it is far better to recognize the speech act by first recognizing the role the sentence plays in the dialogue with the discourse processor since this makes it possible for further processing to take place, such as ellipsis and anaphora resolution.[6]

You will notice that Figure 6 indicates that the biggest difference in terms of speech act recognition between the two mechanisms is that Extended TST got more correct where Standard TST got more acceptable. This is largely because of cases like the one in Figure 4. Sentence 5 is an acceptance to the suggestion made in sentence 3. With Standard TST, the inference chain for sentence 3 would no longer be on the active path when sentence 5 is processed. Therefore, the inference chain for sentence 5 cannot attach to the inference chain for sentence 3. This makes it impossible for the discourse processor to recognize sentence 5 as an acceptance. It will try to attach it to the active path. Since it is a statement informing the listener of the speaker's schedule, a possible speech act is State-Constraint. And any State-Constraint can attach to the active path as a confirmation because the constraints on confirmation attachments are very weak. Since State-Constraint is weaker than Accept, it is counted as acceptable. While this is acceptable for the purposes of speech act recognition, and while it is better than failing completely, it is not the correct discourse structure. If the reply, sentence 5 in this example, contains an abbreviated or anaphoric expression referring to the date and time in question, and if the chain of inference attaches to the wrong place on the plan tree as in this case, the normal procedure for augmenting the shortened referring expression from context could not take place correctly as the attachment is made.

In a separate evaluation with the same set of dialogues, performance in terms of attaching the current chain of inference to the correct place in the plan tree for the purpose of augmenting temporal expressions from context was evaluated. The results were consistent with what would have been expected given the results on speech act recognition. Standard TST achieved 64.3% accuracy while Extended TST achieved 70.4%.

While the results are less than perfect, they indicate that Extended TST outperforms Standard TST on spontaneous scheduling dialogues. In summary, Figure 6 makes clear, with the extended version of TST, the number of speech acts identified correctly increases from 161 (72%) to 171 (77%), and the number of sentences which the discourse processor was able to assign a speech act based on plan inference increases from 164 (74%) to 186 (83%).

---
[6]Ellipsis and anaphora resolution are areas for future development.

## 5 Conclusions and Future Directions

In this paper we have demonstrated one way in which TST is not adequate for describing the structure of discourses with multiple threads in a perspicuous manner. While this study only explores the structure of negotiation dialogues, its results have implications for other types of discourse as well. This study indicates that it is not a structural property of discourse that Attentional State is constrained to exhibit stack like behavior. We intend to extend this research by exploring more fully the implications of our extension to TST in terms of discourse focus more generally. It is clear that it will need to be limited by a model of resource bounds of attentional capacity (Walker 1993) in order to avoid overgenerating.

We have also described our extension to TST in terms of a practical application of it in our implemented discourse processor. We demonstrated that our extension to TST yields an increase in performance in our implementation.

## 6 Acknowledgements

This work was made possible in part by funding from the U.S. Department of Defense.